\documentclass[final,5p,twocolumn]{elsarticle} 
\usepackage{amsmath,amssymb,epsfig}
\usepackage{bm}
\usepackage[ruled]{algorithm2e}
\usepackage{subfigure}
\usepackage{psfrag}
\usepackage{url}
\usepackage[usenames]{color}
\definecolor{orange}{rgb}{0.76,0.22,0.00}
\def\C{{\bf C}}
\def\c{{\bf c}}

\def\f{{\bf f}}

\def\K{{\bf K}}
\def\k{{\bf k}}

\def\m{{\bf m}}

\def\X{{\bf X}}
\def\x{{\bf x}}

\def\y{{\bf y}}

\def\0{{\bf 0}}
\def\1{{\bf 1}}

\def\<{\, \langle \,}
\def\>{\, \rangle \,}
\def\inv{^{-1}}
\def\ts{^\top}

\def\balpha{\bm{\alpha}}

\def\bupsilon{\bm{\upsilon}}

\def\btheta{\bm{\theta}}

\def\DN{\mathcal{N}}

\newcommand{\argmax}{\operatornamewithlimits{argmax}}

\newcommand{\ie}{i.e.\ }
\newcommand{\eg}{e.g.\ }
\newcommand{\wrt}{w.r.t.\ }

\newcommand{\tmp}[1]{{\color{orange} {\bf #1}}}
\newcommand{\quotes}[1]{``#1''}

\def\ctilde{\kern -.04em\lower .7ex\hbox{\~{}}\kern .04em}

\journal{Neurocomputing}
\begin{document}
\begin{frontmatter}
\title{Predictive Active Set Selection Methods for Gaussian Processes}
\author{Ricardo Henao}
\ead{rhenao@binf.ku.dk}
\author{Ole Winther}
\ead{owi@imm.dtu.ku.dk}
\address{ DTU Informatics, Technical University of Denmark, Denmark \\
 		  Bioinformatics Centre, University of Copenhagen, Denmark }
\begin{abstract}
	We propose an active set selection framework for Gaussian process classification for cases when the dataset is large enough to render its inference prohibitive. Our scheme consists of a two step alternating procedure of active set update rules and hyperparameter optimization based upon marginal likelihood maximization. The active set update rules rely on the ability of the predictive distributions of a Gaussian process classifier to estimate the relative contribution of a datapoint when being either included or removed from the model. This means that we can use it to include points with potentially high impact to the classifier decision process while removing those that are less relevant. We introduce two active set rules based on different criteria, the first one prefers a model with interpretable active set parameters whereas the second puts computational complexity first, thus a model with active set parameters that directly control its complexity. We also provide both theoretical and empirical support for our active set selection strategy being a good approximation of a full Gaussian process classifier. Our extensive experiments show that our approach can compete with state-of-the-art classification techniques with reasonable time complexity. Source code publicly available at \url{http://cogsys.imm.dtu.dk/passgp}.
\end{abstract}
\cortext[cor1]{Corresponding author}
\end{frontmatter}
\section{Introduction}
Classification with Gaussian process (GP) priors has many attractive features, for instance it is non-parametric, exceptionally flexible through covariance function designs, provides fully probabilistic outputs and Bayesian model comparison as principled framework for automatic hyperparameter elicitation and variable selection. However, such a set of features comes in with a great disadvantage since the computational cost of performing inference scales cubically with the size, $N$, of the training set. In addition, the memory requirements scale quadratically also with $N$. This means that applicability of Gaussian process classifiers (GPCs) is sadly limited to problems with dataset sizes in the lower ten thousands. The poor scaling of specially non-linear classification methods has inspired a considerable amount of research effort focused on sparse approximations \cite{candela05,rasmussen06,keerthi06,naish08,joachims09,kapoor09,henao10a}. See particularly \cite{candela05,rasmussen06} for a detailed overview of sparse approximations in GPCs. These methods attempt in general to decrease the computational cost of inference in one degree \wrt $N$, \ie $\mathcal{O}(NM^2)$, where $M<N$ and $M$ is the size of a working set consisting on a subset of the training data or a set of auxiliary unobserved variables. Both ways of defining the working set basically target the same objective of getting as close as possible to the classifier that uses the information of the entire training set, however they approach it from different angles. Using a subset from the entire data pool amounts to keep those data points that better contribute to the classification task and discard the remaining ones through some suitable data selection/ranking procedure \cite{lawrence03,seeger03,lawrence05,kapoor09,henao10a}. Alternatively, building an auxiliary set tries to directly reduce the difference in distribution between the classifier using $N$ points and the one using only $M$, by estimating the location of an auxiliary set in the input space, usually called pseudo-input set \cite{candela05,naish08,titsias09}. The latter approach is evidently more principled, however the number of parameters to be learnt grows with the number and size of the auxiliary set, making it unfeasible for datasets in the upper ten thousands and sensitive to overfitting due to the number of free parameters in the model. From a fully Bayesian perspective, in \cite{zhang11} the authors propose an efficient MCMC based inference that is made possible by using a sparse and approximate basis function expansion over the training data set. The main computational burden is therefore the same as other sparse kernel methods possibly with a larger pre-factor due to sampling.

Having in mind that our main goal is to obtain the best classification performance with the least computational cost possible, we do not attempt to estimate auxiliary sets but rather to select a subset of the training data. The framework presented here, Predictive Active Set Selection (PASS-GP) uses the predictive distribution of a GPC in order to quantify the relative importance of each datapoint and then use it to iteratively update an active set. Recently, \cite{kapoor09} proposed a similar criterion in the context of active learning with Gaussian processes. We use the term active set because it is ultimately the one used to estimate the predictive distribution that produces the classification rule and active set updating scheme. In a nutshell, our framework consists of alternating between active set updates and hyperparameter optimization based upon the marginal likelihood of the active set. We provide two active set update schemes that target different practical scenarios. The first simply called PASS-GP builds the active set by including/removing points with small/large predictive probability until no more or too few data points are included in the active set. This means that the size of the active set is not known in advance so as the expected computational complexity. The second scheme is aware that in some applications is very important to keep the computational complexity and/or memory requirements on a budget, thus being able to specify the size of the active set beforehand is essential. In fixed PASS-GP (fPASS-GP) we keep the size of the active set constant by including and removing the same amount of data points in each update to achieve the desired behavior.

The remainder of the paper presents in Section \ref{sec:GPEP} a concise description of expectation propagation based inference for GPCs. Section \ref{sec:our} continues with our proposed framework for active set selection, followed by some theoretical insights based upon a `representer theorem' for the predictive mean of a GP classifier in Section \ref{sec:representer}. Marginal likelihood approximations to the full GP classifier are introduced in section \ref{sec:ML}. Finally, experimental results and discussion appear in Sections \ref{sec:res} and \ref{sec:disc}, respectively.
\section{Gaussian Processes for Classification} \label{sec:GPEP}
Given a set of input random variables $\X=[\x_1,\ldots,\x_N]\ts$, a Gaussian process is defined as a joint Gaussian distribution over function values at the input points $\f=[f_1,\ldots,f_N]\ts$ with mean vector $\m$ (taken to be zero in the following) and covariance matrix $\K$ with elements $K_{ij}=k({\bf x}_i,{\bf x}_j)$ and hyperparameters $\btheta$. For classification, assuming independently observed binary $\pm 1$ labels $\y=[y_1,\ldots,y_N]\ts$ and a probit (cumulative Gaussian) likelihood function $t(y_n|f_n)=\Phi(f_ny_n)$, we end up with an intractable posterior 
\begin{align*}
	p(\f|\X,\y)=Z\inv p(\f|\X)\prod_{n=1}^N t(y_n|f_n) \ ,
\end{align*}
where the normalizing constant $Z=p(\y|\X)$ is the marginal likelihood. If we want to perform inference we must resort to approximations. Here we use Expectation Propagation (EP) because it is currently the most accurate deterministic approximation, see \eg \cite{rasmussen06,kuss05}. In EP, the likelihood function is locally approximated by an un-normalized Gaussian distribution to obtain
\begin{align}
	q(\f|\X,\y) = & \ Z_{\rm EP}\inv p(\f|\X)\prod_{n=1}^N z_n\inv\widetilde{t}(y_n|f_n) \nonumber \\
	= & \ Z_{\rm EP}\inv p(\f|\X)\DN(\f|\widetilde{\m},\widetilde{\C}) \ , \nonumber \\
	= & \ \DN(\f|\m,\c) \ , \label{eq:02}
\end{align}
where $q(\f|\X,\y)\approx p(\f|\X,\y)$, the $z_n$ are the normalization coefficients, $\widetilde{t}(y_n|f_n)$ and $\DN(\f|\widetilde{\m},\widetilde{\C})$ conform the site Gaussian approximations to $t(y_n|f_n)$. In order to obtain $q(\f|\X,\y)$, one starts from $q(\f|\X,\y)=p(\f|\X,\y)$ and update the individual $\widetilde{t}_n$ site approximations sequentially. For this purpose, we delete the site approximation $\tilde{t}_n$ from the current posterior leading to the so called cavity distribution
\begin{align*}
	q_{\backslash n}(\f|\X,\y_{\backslash n}) = p(\f|\X)\prod_{i\neq n} z_i\inv\widetilde{t}(y_i|f_i) \ ,
\end{align*}
from which we can obtain a cavity predictive distribution
\begin{align}
	q_{\backslash n}(y_n|\X,\y_{\backslash n}) = & \ \int t(y_n|f_n) q_{\backslash n}(\f|\X,\y_{\backslash n}) d\f \ , \nonumber \\
	= & \ \Phi\left(\frac{y_n m_{\backslash n}}{\sqrt{1+v_{\backslash n}}}\right) \ , \label{eq:03}
\end{align}
where $m_{\backslash n}=v_{\backslash n}(C_{nn}\inv m_n-\widetilde{C}_{nn}\inv\widetilde{m}_n)$ and $v_{\backslash n}=(C_{nn}\inv-\widetilde{C}_{nn}\inv)\inv$. We then combine the cavity distribution with the exact likelihood $t(y_n|f_n)$, to obtain the so called tilted distribution $q_n(\f|\X,\y)=z_n\inv t(y_n|f_n)q_{\backslash n}(\f|\X,\y_{\backslash n})$. Since we need to choose the parameters of the site approximations we must minimize some divergence measure. It is well known that when $q(\f|\X,\y)$ is Gaussian, minimizing $\mathrm{KL}(p(\f)||q(\f))$ is equivalent to moment matching between those two distributions including zero-th moments for the normalizing constants. The EP algorithm iterates by updating each site approximation in turn and makes several passes over the training data.

With the Gaussian approximation to the posterior distribution in equation \eqref{eq:02}, it is possible to calculate the predictive distribution of a new datapoint $\x^\star$ as
\begin{align}
	q(y^*|\X, \y,\x^\star) = & \int t(y^\star|f^\star)q(f^\star|\X,\y,\x^\star)df^
	\star \ ,\nonumber \\
	= & \ \Phi\left(\frac{y^\star m^\star}{\sqrt{1+v^\star}}\right) \ , \label{eq:04}
\end{align}
where $q(f^\star|\X,\y,\x^\star)$ is the approximate predictive Gaussian distribution (the marginal of $q(\f,f^\star|\X,\y,\x^\star)$ \wrt $\f$) with mean $m^\star=\k^{\star\top}(\K+\widetilde{\C})\inv\widetilde{\m}$ and variance $v^\star=k^{\star\star}-\k^{\star\top}(\K+\widetilde{\C})\inv\k^\star$. In addition, the approximation to the marginal likelihood $p(\y|\X)$ results in the normalization constant from equation \eqref{eq:02}, \ie $q(\y|\X)=Z_{\rm EP}$. The logarithm of $Z_{\rm EP}(\btheta,\X,\y)$ and its derivatives can be used jointly with conjugate gradient updates to perform model selection under the evidence maximization framework. For a detailed presentation of GP including its implementation details, consult \cite{rasmussen06,kuss05}.
\section{Predictive Active Set Selection} \label{sec:our}
The EP algorithm is performed by iterative updates of each site approximation using the whole dataset $\{\X,\y\}$. In the active set scenario on the other hand, we only want to approximate the posterior distribution in equation \eqref{eq:02} using a small subset, the active set $\{\X_A,\y_A\}$. Since exploring all possible active sets is obviously intractable even for a fixed active set size $M$, the problem is how to select an active set that delivers a performance as good as possible within the available computing resources. The Informative Vector machine (IVM) \cite{lawrence03} for instance, computes in each iteration the differential entropy score for all data points not already part of the active set $\{\X_I,\y_I\}$ and perform updates by including the single point leading to a maximum score. Despite this greedy heuristic, IVM has proved to behave quite well in practice, giving the so far best reported GP performance on the USPS and MNIST tasks \cite{lawrence03,seeger03}. We propose an iterative approach in the same spirit with two main conceptual changes:
\vspace{3mm}
\begin{list}{\labelitemi}{\leftmargin=1em \topsep=0em \parsep=0em}
	\item {\bf Active set inclusion/deletion} based directly upon the data point weight in prediction. The `representer theorem' for the mean prediction, discussed in Section \ref{sec:representer}, leads directly to the weight being expressed in terms of (a derivative of) the cavity predictive probability. This means that we can actually use the predictive distribution for a point in the inactive set to predict the weight it would have if it would be included in the active set. For classification we use the (cavity) predictive probability to decide upon deletion and inclusion because it is monotonically related to weight and it is a readily interpretable quantity.
	\item {\bf Hyperparameter optimization} must be an integral part of algorithm, because the weights of the examples (and thus the active set) is conditioned on the hyperparameter values and vice versa. We therefore alternate between active set updates and hyperparameter optimization using several passes over the data set.
\end{list}
\vspace{3mm}
Next we discuss the details of our (f)PASS-GP framework followed by a detailed comparison with the IVM. First we need to define rules for including and deleting points of the active set. As already mentioned, we use the predictive distribution in equation \eqref{eq:04} for inclusions since data points with small predictive probability are more likely to contribute to improve the classifier performance and the quality of the active set. For deletions, we use the cavity predictive distribution in equation \eqref{eq:03} because when examined carefully it can be seen as a leave-one-out estimator \cite{opper00}. This means that points with cavity probability close to one do not contribute to the decision rule thus they can be discarded from the active set. With the two ranking measures set, \ie equations \eqref{eq:03} and \eqref{eq:04}, we have essentially two possibilities. The first is to set probability thresholds on the distributions and let the model decide the size of the active set or we can rather specify directly the amount of inclusions/deletions. In PASS-GP, we include points in the active set with probability less than $p_{\rm inc}$ and remove them with probability greater than $p_{\rm del}$. The appealing aspect of these thresholds is that they can be interpreted, for instance if we set $p_{\rm inc}=0.5$ we will include all misclassified observations in the current active set whereas if $p_{\rm inc}=0.6$ we will also include points near the decision boundary. We require two thresholds because we only want to remove points that as for the classifier are very easy to classify, so unlike $p_{\rm inc}$, $p_{\rm del}$ must be close enough to one. In fPASS-GP, we want to keep the computational complexity of the classifier under control thereby we want the size of the active set to be fixed. For this purpose we only have to be sure that each active set update includes and removes the same amount of points. In practice we define $p_{\rm exc}$ as the exchange proportion \wrt $M$, meaning that each update replaces the fixed proportion of most hard to classify points in the inactive set with those more surely classified in the current active set. This update rule assumes that the active set is large enough to contain points in the active set with cavity probability close to one.

From a practical point of view, ranking every point in the inactive set at each iteration for inclusion could become prohibitive for large datasets. However we still want to be able to cover the whole dataset rather than selecting a random subset for ranking. We then split the data into $N_{\rm sub}$ non-overlapping subsets and process each one of them in each iteration, such that each batch has something between 100 and 1000 data points. 

Hyperparameter selection is a very important feature and needs to be done jointly with the active set update procedure. Algorithm \ref{alg:01} starts from a fixed randomly selected active set of size $N_{\rm init}$ (that is $M$ in fPASS-GP), large enough to provide a good initial hyperparameter set values. Next we alternate between active set and hyperparameter optimization updates. Having in mind that we only expect small changes of the hyperparameters from one iteration to another, we reuse current values of $\btheta$ as initial values for the next iteration to speed-up the learning process. The addition and deletion rules in Algorithm \ref{alg:01} have parameters $\{p_{\rm inc},p_{\rm del}\}$ and $p_{\rm exc}$ for PASS-GP and fPASS-GP, respectively.

\IncMargin{0.5mm}
\begin{algorithm}[t]
	\DontPrintSemicolon
	\SetKwData{remove}{RemoveRule($q_{\backslash n}(y_n|\X_A,\y_{A\backslash n})$)}
	\SetKwData{addition}{AdditionRule($q(y^*|\X_A,\y_A,\x^\star,\bupsilon)$)}
	\SetKwData{SplitData}{SplitData}
	\SetKwData{TuneParams}{TuneParams}
	\SetKwData{RunEP}{RunEP}
	\SetKwData{Predict}{Predict}
	\SetKwInOut{Input}{Input}
	\SetKwInOut{Output}{Output}
	\Input{$\{\X,\y\}$, $\btheta$ and $\{N_{\rm init}$, $N_{\rm sub}$, $N_{\rm pass}\}$}
	\Input{$p_{\rm inc}$ and $p_{\rm del}$ (PASS-GP)}
	\Input{$p_{\rm exc}$ (fPASS-GP)}
	\Output{$q(\f_A|\X_A,\y_A)$, $\bm{\theta}_\mathrm{new}$ and $A$}
	\Begin{
		$A\leftarrow\{1,\ldots,N_{\rm init}\}$ \;
		$\{\X,\y\}_{\rm sub}^{(1)},\ldots\{\X,\y\}_{\rm sub}^{(N_{\rm sub})}\leftarrow\{\X,\y\}$ \;
		\For{$i=1$ \KwTo $N_{\rm pass}$}{
			\For{$j=1$ \KwTo $N_{\rm sub}$}{ 
				$\btheta_{\rm new} = \argmax_{\btheta} \log Z_{\rm EP}(\btheta,\X_A,\y_A)$ \;
				Get $q(\f_A|\X_A,\y_A)$ and $q(y^*|\X_A,\y_A,\x^\star)$ \;
				\ForAll{$\{\x_n,y_n\}\in\{\X_A,\y_A\}$}{
					\lIf{\remove}{ $A\leftarrow A\backslash
\{n\}$}
				}
				\ForAll{$\{\x_n,y_n\}\in\{\X,\y\}_{\rm sub}^{(j)} $}{
					\lIf{\addition}{ $A\leftarrow A\cup \{n\}$ }
				} 
			}
		} 
	}
	\caption{Predictive active set selection}
	\label{alg:01}
\end{algorithm}
\DecMargin{0.5mm}
\subsection{Differences between (f)PASS-GP and IVM}
Since IVM is the closest relative of our active set selection method, we briefly discuss the main differences between the two: (i) The active set and thus the computational complexity is usually fixed beforehand in IVM. PASS-GP works with inclusion and deletion thresholds instead. (ii) IVM does not allow for deletions from the active set which is a clear disadvantage as points often become irrelevant at a later stage, when more points have been included. In (f)PASS-GP we can make an (almost) unbiased common ranking of all training points active as well as inactive, using a quantity that is meaningful and directly related to the weight of the training point in predictions. Using both inclusions/deletions and several passes over the training set makes (f)PASS-GP quite insensitive to the initial choice of active set. (iii) When the dataset is considerably large, IVM randomly selects a subset of points to be ranked from the inactive set, meaning that is likely that some points of the dataset are never considered for inclusion in the active set. (iv) The hyperparameter optimization is a part of the algorithm in (f)PASS-GP working on subsets of data between updates and iterating over the data set several times. IVM makes a single inclusion per step and in principle stops when the limit for the active set is reached. (iv) In terms of complexity time per iteration IVM is faster than (f)PASS-GP, $\mathcal{O}(NM)$ against $\mathcal{O}(M^2(2+N/N_{\rm sub}))$ where $M$ is the size of $A$, however storage requirements are considerable lower, $\mathcal{O}(M^2)$ compared to $\mathcal{O}(NM)$.
\section{Representer for Mean Prediction}\label{sec:representer}
The `representer theorem' for the posterior mean of $\f$ \cite{opper00}, connects the predictive probability and the weight of a data point. Using that $p(\f|\X) = -\K \frac{\partial}{\partial \f}p(\f|\X)$, we get the exact relation for the posterior mean $\<\f\> = \K\balpha$ with the weight of element $n$ being
\begin{align*}
	\alpha_n = & \ \frac{1}{p(\y|\X)} \int p(\f|\X)\frac{\partial}{\partial f_n} p(\y|\f) d\f \\
	= & \ \frac{ \< p'(y_n|f_n)\>_{\backslash n} }{\< p(y_n|f_n) \>_{\backslash n} } 
	= \left.\frac{\partial}{\partial h} \log \< p(y_n|f_n+h) \>_{\backslash n}\right|_{h=0} \ ,
\end{align*}
where $\<\cdot\>_{\backslash n}=m_{\backslash n}$ denotes an average over a posterior without the $n$-th data point and $p'(y_n|f_n) = \partial p(y_n|f_n) / \partial f_n$. The final expression implies that the weight is nothing but the log derivative of the cavity predictive probability $\<p(y_n|f_n)\>_{\backslash n}$ $=p(y_n|\X,\y_{\backslash n})$. For regression, $p(y_n|f_n) = \DN(y_n|f_n,\sigma^2)$ and $\alpha_n = (y_n-\< f_n \>_{\backslash n})(\sigma^2+v_{\backslash n})\inv$ with $v_{\backslash n} =\<f_n^2\>_{\backslash n} - \<f_n\>^2_{\backslash n}$. The element $\alpha_n$ will therefore be small when the cavity mean has a small deviation from the target relative to the variance. For a new data point pair $\{\x^\star,y^\star\}$, we can calculate the weight of this point \emph{exactly}, replacing the cavity average with the full average in the expression above. We can therefore predict without any EP rerunning, how much weight this new point will have. For classification we can calculate the weight using the current EP approximation. When $z_n= y_n \<f_n\>_{\backslash n}/\sqrt{1+v_{\backslash n}}$ is above $\approx 4$, the cavity probability equation \eqref{eq:03} approaches one and $\alpha_n \approx y_n \exp(-z_n^2/2)/\sqrt{2\pi (1+v_{\backslash n})}$. This fast decay indicates that GPC in many cases will be effectively sparse even though $\balpha$ strictly does not contain zeros.

In the inclusion/deletion steps we rank data points according to their weights. For classification we can indeed use the predictive probability directly, since it is a monotonic function of the weight. Including a new data point will of course affect the value of all other weights as well leading to a rearrangement of their rank. Including multiple data points will also invalidate the predicted value of the weights (e.g.\ think of the extreme of two new data points being identical). We therefore have to recalculate the weights by retraining with EP for classification or simply updating the posterior for regression before going to the next step. If we have already an active set covering the decision regions well enough, this rearrangement step will amount to minor adjustments and the approximation will work well.

In this work we have only used the representer theorem for active set selection. It is also possible, but not tested here, to use all training points for prediction while only calculating the posterior on the active set. The inactive set weights are then simply set to the predicted values from the active set posterior. To get the full predictive probability one also has to calculate the contribution to the predictive variances which can be obtained by a similar theorem but for the predictive variance, see \cite{opper00}.
\section{Marginal Likelihood Approximations} \label{sec:ML}
In this section we decompose the marginal likelihood in their active and inactive set contributions. We will argue that the contribution from the active set will dominate, justifying why we can limit ourselves to optimizing the hyperparameters over this set. In the following section we will investigate this assumption empirically. The marginal likelihood can be decomposed via the chain rule as
\begin{align}
	p(\y|\X) = \ p(\y_I|\y_A,\X_A,\X_I) p(\y_A|\X_A) \ , \label{eq:05}
\end{align}
where we have used the marginalization property of GPs, 
\begin{align*}
	p({\bf y}_A|{\bf X})= \int p({\bf y}_A|{\bf f}_A)p({\bf f}_A|{\bf X}_A)d{\bf f}_A =
	p({\bf y}_A|{\bf X}_A) \ ,
\end{align*}
that we approximate as $q(\y_A|\X_A)=Z_{\mathrm{EP},A}$ and we identify it as the marginal likelihood for the active set $A$. The conditional marginal likelihood term can be written as
\begin{align}
	& p(\y_I|\y_A,\X_A,\X_I) = \nonumber \\
	& \hspace{4mm} \int p(\y_I|\f_I) p(\f_I|\X_I,\X_A,\f_A) p(\f_A|\X_A,\y_A) \, d\f_A d\f_I \ , \label{eq:06}
\end{align}
where we used $p(\f|\X) = p(\f_I|\X_I,\X_A,\f_A)p(\f_A|\X_A,\y_A)$. We can make an EP approximation here just like in equation \eqref{eq:02} by replacing the posterior $p(\f_A|\X_A,\y_A)$ by the multivariate Gaussian $q(\f_A|\X_A,\y_A) = \DN(\f_A|\m_A,\C_{AA})$ where active set specific means and variances are found by EP. Marginalizing over $\f_A$ in equation \eqref{eq:06} makes it now tractable
\begin{align*}
	q(\y_I|\y_A,\X_A,\X_I) \approx \int p(\y_I|\f_I) \DN(\f_I|\m_{I|A},\C_{II|A}) d\f_I \ ,
\end{align*}
with parameters
\begin{align*}
	{\bf m}_{I|A} = & \ {\bf K}_{IA}( {\bf K}_{AA} + \widetilde{{\bf C}}_{AA})^{-1} \widetilde{{\bf m}}_A \ , \\
	{\bf C}_{II|A} = & \ {\bf K}_{II} - {\bf K}_{IA}( {\bf K}_{AA} + \widetilde{{\bf C}}_{AA})^{-1} {\bf K}_{AI} \ ,
\end{align*}
where the tilted moments are as defined in Section \ref{sec:GPEP}. When the inactive set consists of a single example, we obtain the EP predictive distribution in equation \eqref{eq:04}, otherwise we have to solve for a new marginal likelihood. Denoting the marginal likelihood for a set $\{\X,\y\}$ with a non-zero mean GP prior by 
\begin{align*}
	Z(\btheta,\X,\y,\m) = \int p(\y|\f) \DN(\f|\m,\K) \, d\f \ ,
\end{align*}
and its EP approximation by $Z_{\rm EP}(\btheta,\X,\y,\m)$, we can write the approximation to the marginal likelihood in equation \eqref{eq:05} as
\begin{equation*} 
Z_{\mathrm{ACC}} \equiv Z_\mathrm{EP}(\btheta,\X,\y_I,\m_{I|A}) Z_{\rm EP}(\btheta,\X,\y_A,\0) \ .
\end{equation*}
Using this approximate decomposition reduces the complexity of EP from ${\cal O}(N^3N_{\rm pass})$ to ${\cal O}((|I|^3+M^3)N_{\rm pass})$, where $|I|$ is the size of the inactive set. Unfortunately this is still too costly for large $N$. A final low complexity approximation to the marginal likelihood, that we denote by $Z_{\mathrm{APP}}$, is to replace $p(y_I|y_A,\X)$ with the product of marginals $\prod_{i\in I} p(y_i|\y_A,\X_A,\x_i)$. Empirically---see Figure \ref{fg:mla}, this approximation turns out to be lower than the actual marginal likelihood, i.e.\ the joint distribution enforces the labels relative to the product of the marginals.
\section{Experiments} \label{sec:res}
The results presented in this section consist of several classification tasks performed on three well known datasets, namely USPS, MNIST and IJCNN. The first two correspond to handwritten digit databases while the third is a physical system inspired dataset assembled for the IJCNN 2001 neural network competition. We compare the two approaches introduced in section \ref{sec:our} against the IVM and Reduced complexity SVM (RSVM) \cite{keerthi06}. We consider as performance measures not only classification errors, but the error-cost trade-off and prediction uncertainty. We also present results for the approximation to the marginal likelihood of the full GP presented in section \ref{sec:ML}. All experiments were performed on a 2.0GHz desktop machine with 2GB RAM.
\subsection{USPS}
The USPS digits database contains 9289 grayscale images of size $16\times16$ pixels, scaled and translated to fall within the range from $-1$ to $1$. Here we adopt the traditional data splitting, i.e. 7291 observations for training and the remaining 2007 for testing. For each binary one-against-rest classifier we use the same model setup consisting of a squared exponential covariance matrix plus additive jitter
\begin{align} \label{eq:rbf}
	k(\x_i,\x_j) = \theta_1\exp\left(-\frac{\|\x_i-\x_j\|^2}{2\theta_2}\right) + \theta_3\delta_{ij}
\end{align}
where $\delta_{ij}=1$ if $i=j$ and zero otherwise. We have three hyperparameters in $\btheta$, namely, signal variance, characteristic length scale and jitter coefficient. Provided that the four active set methods being considered may depend upon random initialization we repeated all tasks 10 times. Individual settings for each method are:
\vspace{4mm}
\begin{list}{\labelitemi}{\leftmargin=1em \topsep=0em \parsep=0em}
	\item { PASS-GP:} $N_\mathrm{init}=300$, $N_\mathrm{sub}=10$, $N_\mathrm{pass}=2$, $p_{inc}=0.6$ and $p_{del}=0.99$.
	\item { fPASS-GP:} $N_\mathrm{init}=300$, $N_\mathrm{sub}=10$, $N_\mathrm{pass}=4$, $p_{exc}=0.02$. We allow fPASS-GP to perform more passes through the data because fPASS-GP progresses slower due to $p_{exc}$ being small.
	\item { RVM:} $M=500$, $\btheta=[1 \ 1/16 \ 0]$, $C=10$ and $\kappa=10$. More precisely, $\btheta$ and the regularization parameter, $C$, were obtained by grid search cross-validation, while $\kappa$ was set to the value suggested by the authors of \cite{keerthi06}.
	\item { IVM:} $M=300$ and $N_\mathrm{pass}=8$. In the publicly available version of IVM, hyperparameter selection is done by alternating between full active set selection and hyperparameter optimization. Since IVM starts from an empty active set, it can be very sensitive to the initial values of $\btheta$. We experienced however that by adding a linear term, $\theta_4\x_i\ts\x_j$, in the covariance matrix in equation \eqref{eq:rbf} makes IVM quite insensitive to initialization. The results reported here include such a linear term because we found that using equation \eqref{eq:rbf} alone makes the IVM to perform very poorly.
\end{list}
\vspace{4mm}

\begin{figure}[ht]
	\centering
		\centering
		\begin{psfrags}
			\psfrag{dig}[c][c][0.6][0]{Digit}
			\psfrag{act}[c][c][0.6][0]{Active set size}
			\psfrag{err}[c][c][0.6][0]{Error ($\%$)}
			\psfrag{pass-gp}[c][c][0.5][0]{PASS-GP}
			\psfrag{fpass-gp}[c][c][0.5][0]{fPASS-GP}
			\psfrag{rsvm}[c][c][0.5][0]{\hspace{1mm} RSVM}
			\psfrag{ivm}[c][c][0.5][0]{\hspace{1mm} IVM}
			\psfrag{fullGPC}[c][c][0.5][0]{full GPC}
			\subfigure[]{\includegraphics[scale=0.4]{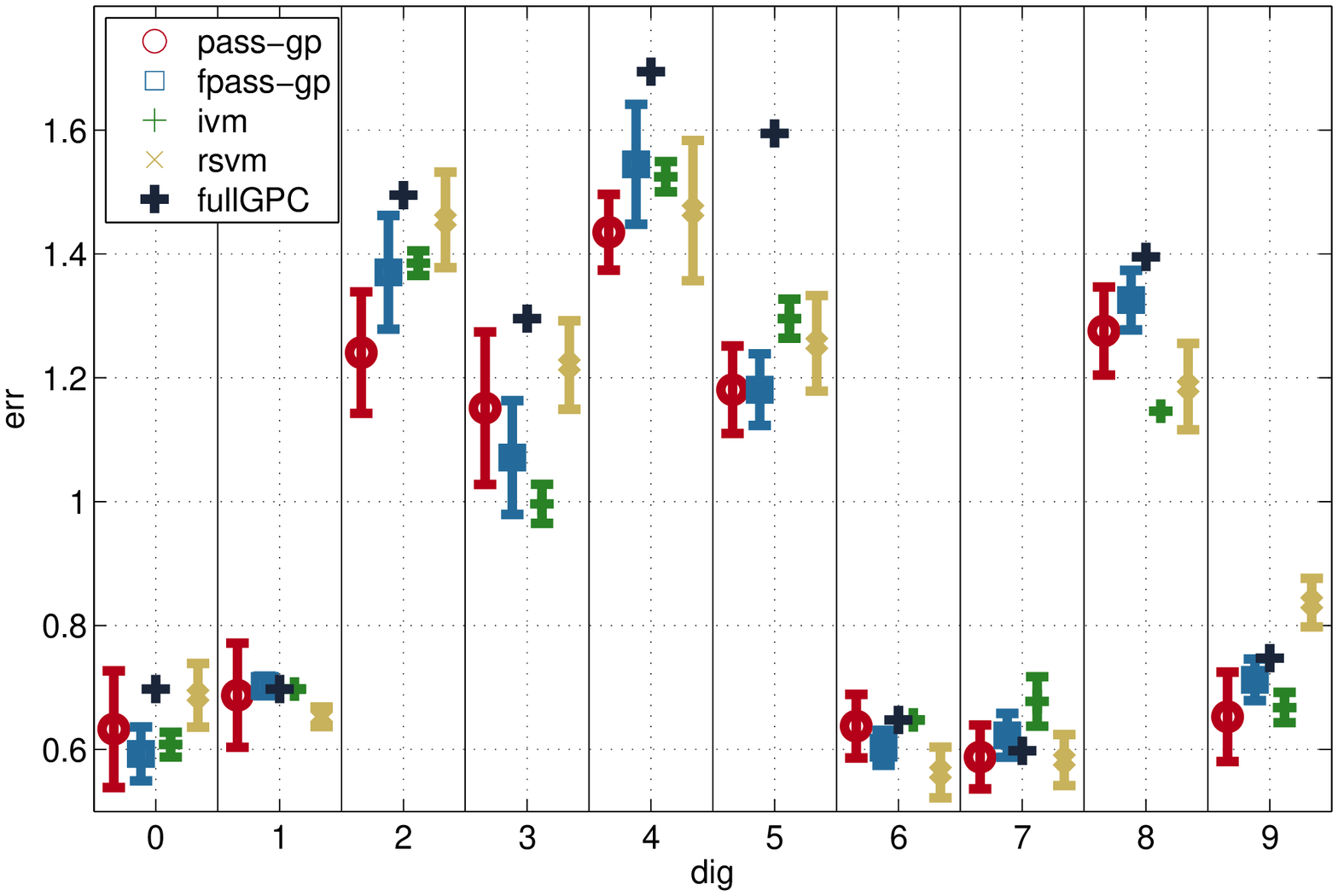}\label{fg:bars}}
			\subfigure[]{\includegraphics[scale=0.4]{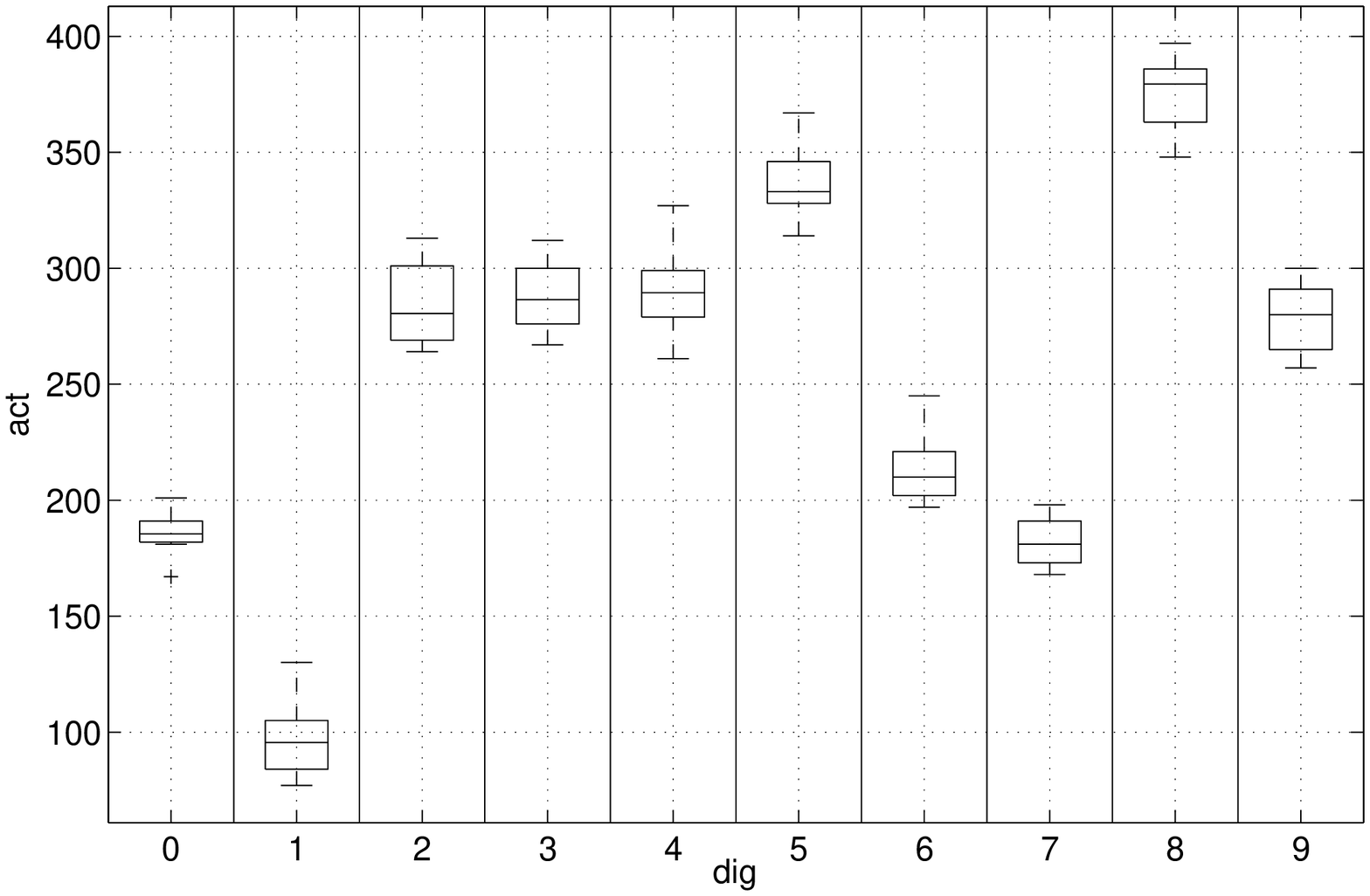}\label{fg:boxes}}
		\end{psfrags}
	\caption{Error rates and active set sizes for USPS data. (a) mean classification errors for each digit using PASS-GP, fPASS-GP, RSVM, IVM and the full GPC with hyperparameter optimization. (b) Active set sizes for PASS-GP. Note that fPASS-GP and IVM use $M=300$, whereas RSVM uses $M=500$ for the results in (a). Error bars are standard deviations over 10 repetitions.}
\end{figure}
Figure \ref{fg:bars} shows mean test errors for every one-against-rest task using PASS-GP, fPASS-GP, RSVM, IVM and the full GPC with hyperparameter optimization. Besides, Figure \ref{fg:boxes} shows the active set sizes for each digit using PASS-GP. From the figure, it can be seen that Gaussian process based active set methods perform similarly still slightly better than the RVM. The full GPC was only ran once due to its computational requirements, which explains the lack of error bars in Figure \ref{fg:bars}. Furthermore, compared to fPASS-GP, IVM ($M=300$) and RSVM ($M=500$), PASS-GP seems to require smaller active sets to achieve similar classification performance. It is important to mention that we also tried larger values of $M$ for the fixed active set algorithms but without any significant improvement in performance. 

\begin{figure}[ht]
	\centering
		\centering
		\begin{psfrags}
			\psfrag{err}[c][c][0.6][0]{Error ($\%$)}
			\psfrag{act}[c][c][0.6][0]{Active set size}
			\psfrag{time}[c][c][0.6][0]{Time (s)}
			\psfrag{pass-gp}[c][c][0.4][0]{PASS-GP}
			\psfrag{fpass-gp}[c][c][0.4][0]{fPASS-GP}
			\psfrag{ssvm}[c][c][0.4][0]{\hspace{1mm} RSVM}
			\psfrag{ivm}[c][c][0.4][0]{\hspace{1mm} IVM}
			\psfrag{fullGPC}[c][c][0.4][0]{\hspace{0.5mm} full GPC}
			\subfigure[]{\includegraphics[scale=0.32]{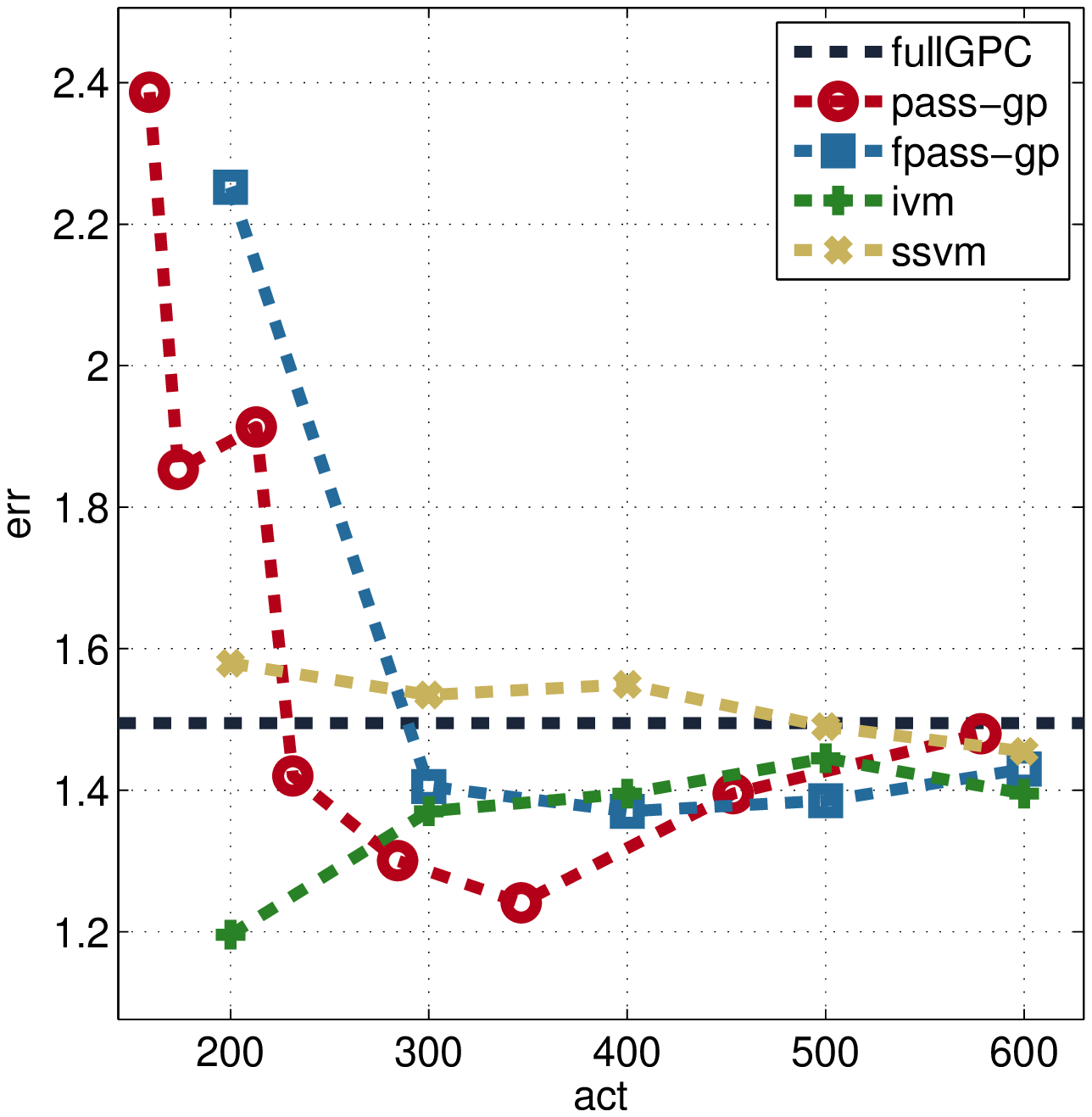}\label{fg:dige3}}
			\subfigure[]{\includegraphics[scale=0.32]{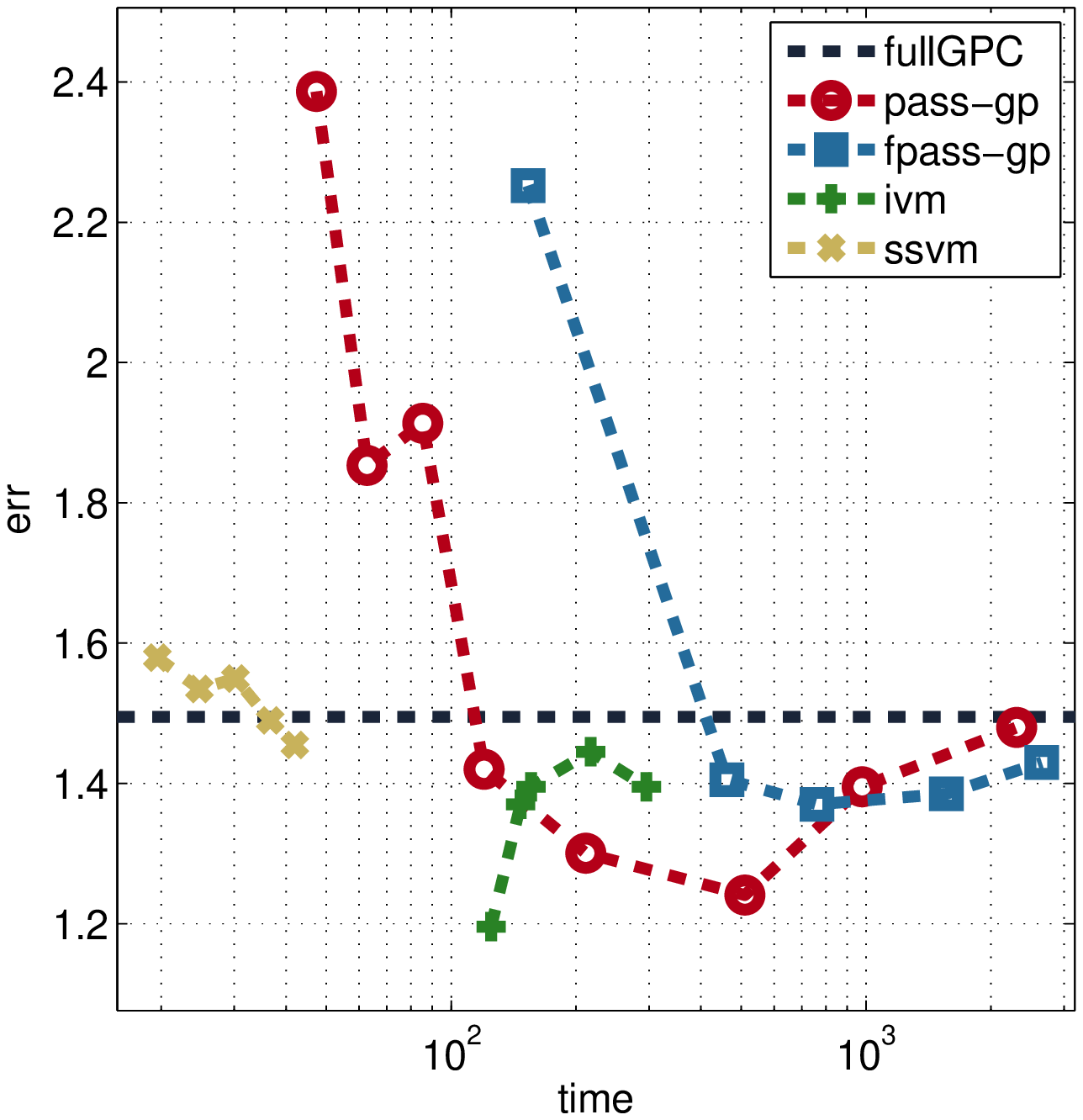}\label{fg:digt3}}
			\subfigure[]{\includegraphics[scale=0.32]{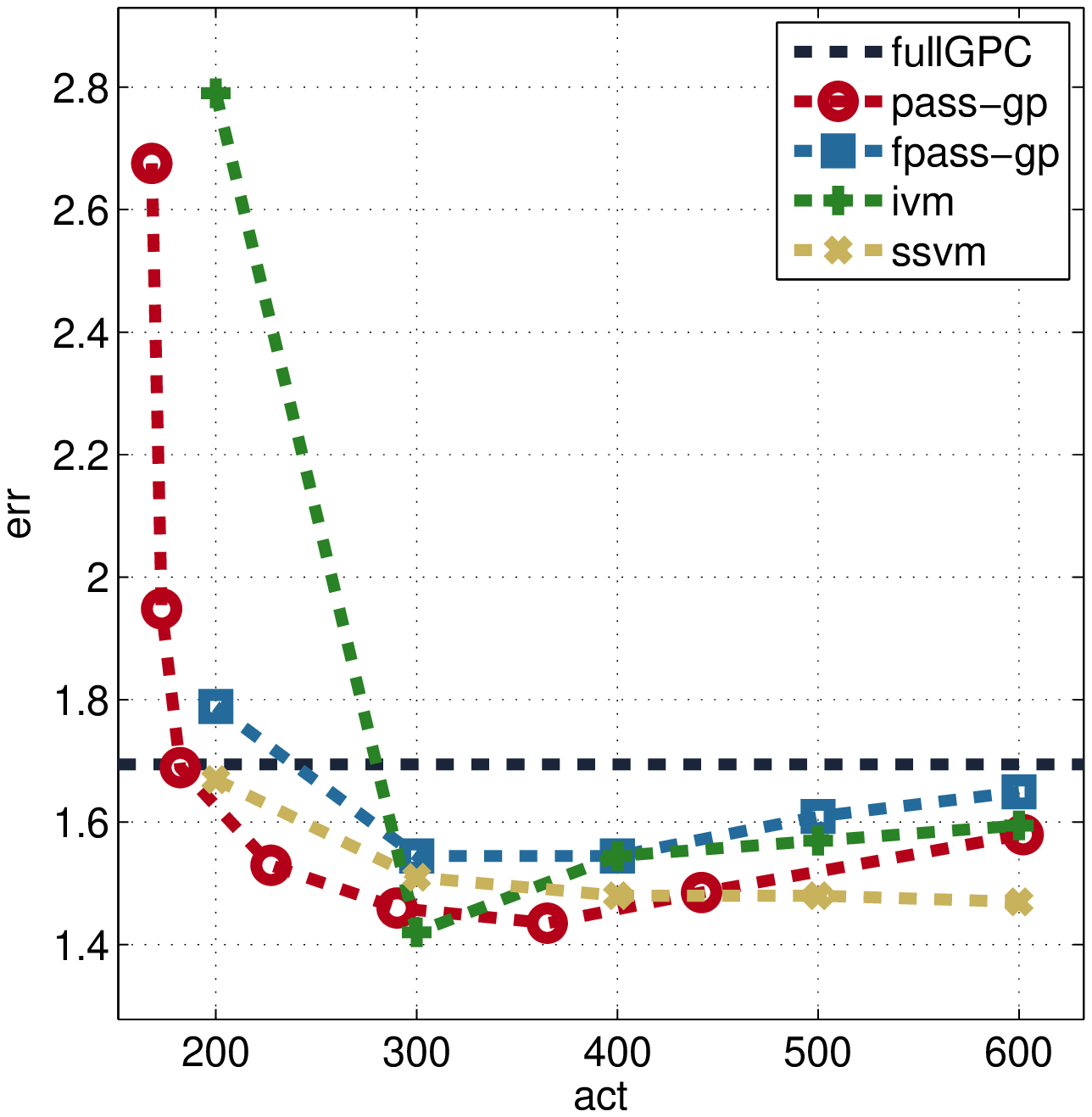}\label{fg:dige4}}
			\subfigure[]{\includegraphics[scale=0.32]{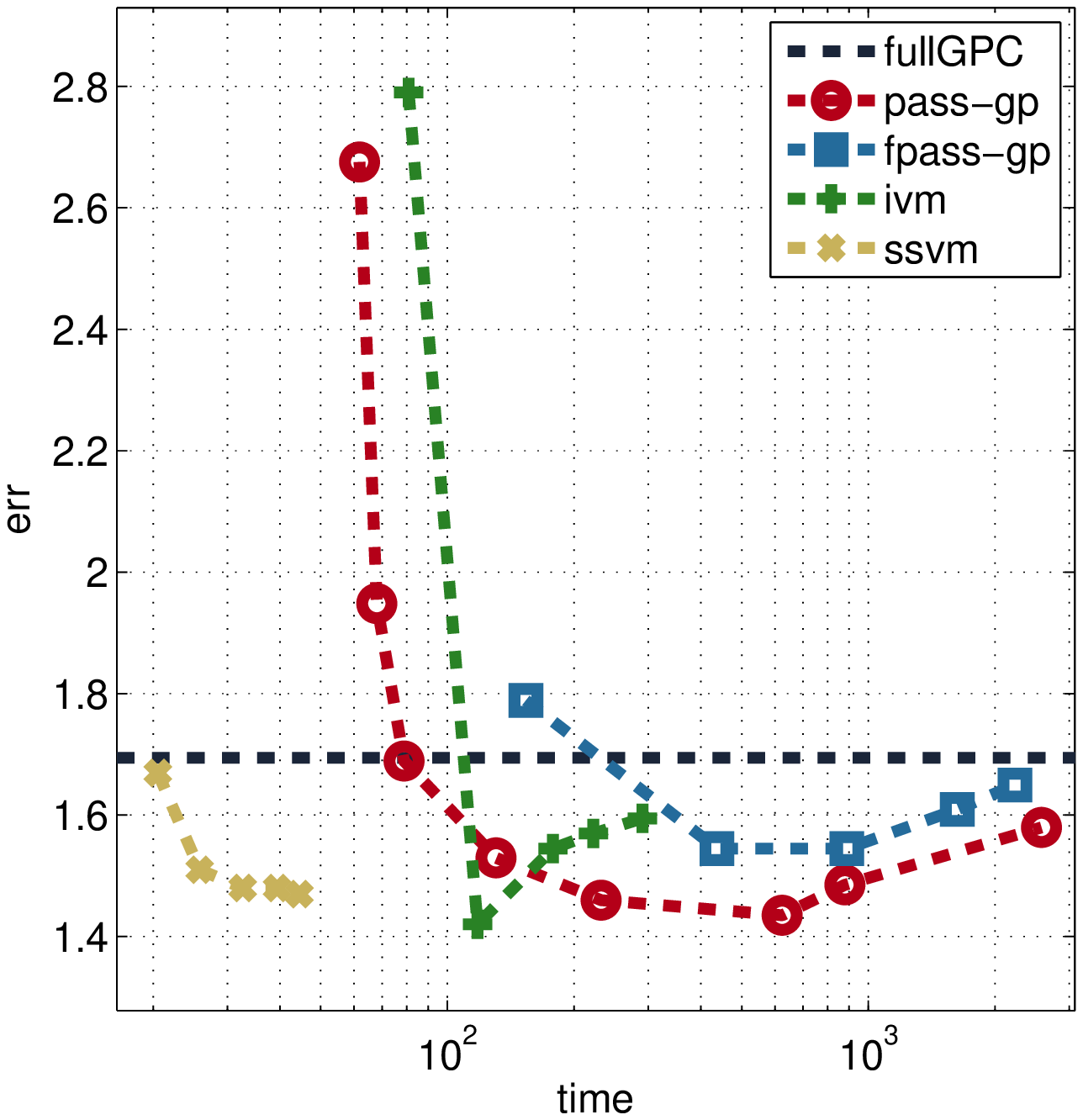}\label{fg:digt4}}
		\end{psfrags}
	\caption{Results for selected individual digits of USPS data. (a) and (c) show mean classification errors as a function of the active set size for digits 2 and 4 vs the rest, respectively. (b) and (c) show mean classification errors as a function of the running time, matching panels (a) and (c). The horizontal dashed line in all plots is the performance of the full GPC with hyperparameter selection. In both cases, the full GPC took approximately $8.6e5$ seconds \cite{henao10a}. Values represent averages over ten independent repetitions with error bars omitted for clarity.}
	\label{fg:dig}
\end{figure}
Figure \ref{fg:dig} show classification errors for digits 2 and 4 against the others in top and bottom panels, respectively, as a function both of the active set size and running time. For fPASS-GP, RSVM and IVM we used $M=\{200,\ldots,600\}$ and for PASS-GP we used $p_{inc}=\{0.2.0.3,\ldots,0.9\}$. We included also the classification error obtained by the full GPC with hyperparameter optimization depicted as an horizontal dashed line. See \cite{henao10a} for a more detailed comparison between PASS-GP and full GPCs. Several features from Figure \ref{fg:dig} worth to be highlighted. (i) Gaussian process based methods approach the full GP for large values of $M$, as expected. (ii) Similar to Figure \ref{fg:bars}, PASS-GP seems to consistently outperform fPASS-GP for similar sizes of $M$. (iii) For small values of $M$, RSVM and IVM perform better than our active set methods, however further increasing $M$ does not considerably improves their performance. When $M$ is small enough, it is very likely that our approaches are not able to obtain plausible estimates of the hyperparameters of the covariance function, thus its poor performance compared to RSVM that uses fixed values. Provided that the full GPC takes $8.6e5$ seconds to run, PASS-GP and fPASS-GP are approximately three orders of magnitude faster than the full GPC with hyperparameter optimization, see \cite{henao10a}. Form Figures \ref{fg:digt3} and \ref{fg:digt4}, we see that for similar active set sizes, PASS-GP and fPASS-GP have comparable computational costs as one may expect. Similarly, RSVM and IVM scale better than our active set selection methods. In terms of error-cost trade-off, RVM has a clear edge while the Gaussian process based methods can be regarded as comparable. It is important to note that for RVM, the difference in computational costs as seen in Figures \ref{fg:digt3} and \ref{fg:digt4} should not be considered as significant since we are not counting the time used to obtain the parameters used by the RSVM, that unfortunately need to be selected by expensive grid search with cross-validation. The IVM turned out to be time-wise comparable to our active set methods not because its selection procedure but due to the hyperparameter optimization scheme used.
\begin{figure*}[th]
	\centering
	\begin{psfrags}
		\psfrag{time}[c][c][0.7]{Time (s)}
		\psfrag{nlz}[c][c][0.7]{$\log Z$}
		\psfrag{ZEP}[c][c][0.7]{$Z_{\rm EP}$}
		\psfrag{ZAPP}[c][c][0.7]{$Z_{\rm APP}$}
		\psfrag{ZACC}[c][c][0.7]{$Z_{\rm ACC}$}
		\psfrag{ZEPA}[c][c][0.7]{$Z_{\rm EP,A}$}
		\psfrag{act}[c][c][0.7]{Active set size}
		\includegraphics[scale=0.51]{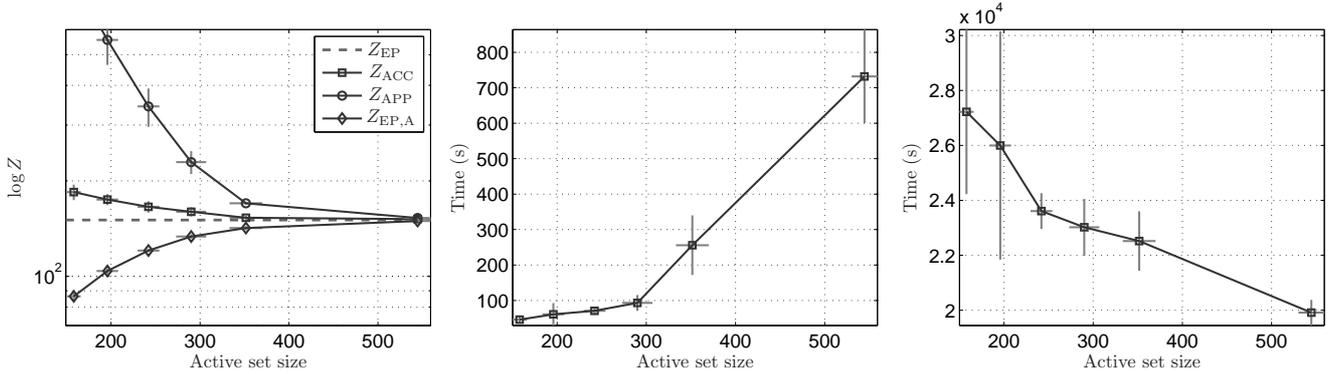}
	\end{psfrags}
	\caption{Marginal log-likelihood approximations as a function of the active set size for digit 3 vs the rest. The plots show means and standard deviations (error bars) over ten repetitions. Each marker indicates a different inclusion threshold $p_{\rm inc}=\{0.5,0.6,\ldots,0.9,0.99\}$. In the left panel, $Z_{\rm EP}$ is for the full GPC ($p_{\rm inc}=1$), $Z_{\rm EP,A}$ for the active set only and the remaining two, $Z_{\rm ACC}$ and $Z_{\rm APP}$, are the proposed approximations. The middle and right panels shows computation times required to compute $\{Z_{\rm APP},Z_{\rm EP,A}\}$ and $Z_{\rm ACC}$, respectively.}
	\label{fg:mla}
\end{figure*}

The results obtained on USPS suggest that (f)PASS-GP is performing slightly better than the full GPC. This could be due to numerical instability produced by the size of the problem, by the iterative nature of the EP algorithm and/or not enough iterations for the hyperparameter selection procedure. However, it could also mean that optimizing on the active set achieves a better \quotes{local} fit around the decision boundary region. A priori, one cannot expect that a single set of hyperparameters is able to describe all regions in input space, thus every possible active set. The same kind of local improvement observed here was also reported by \cite{snelson06} and \cite{naish08} using GPC auxiliary set methods.

Combining the ten binary tasks into a one-against-rest multi-class classifier, PASS-GP obtained $4.51\pm0.17\%$ which is comparable or better\footnote{Assuming independent errors, the standard deviation on the performance is $\sqrt{\epsilon(1-\epsilon)/N_{\rm test}}$ giving approximately $0.4\%$ for USPS and $0.1\%$ for MNIST.} than $4.61\pm0.11\%$ by fPASS-GP, $4.88\pm0.12\%$ by RSVM and $4.38\pm0.11\%$ by IVM. Baselines are, $5.13\%$ by GPC with hyperparameter optimization, $4.78\%$ by GPC with fixed $\btheta$ and $9.75\pm0.40\%$ by GPC with random active set selection. Other relevant results found in the literature include $5.15\%$ by online GP \cite{csato02} and $4.98\%$ by IVM with randomized greedy selection \cite{seeger03}. All three Gaussian process based methods are comparable with state-of-the-art techniques such as SVM, see \cite{smola01}. It is worth pointing out that the best result we could obtain from IVM using the covariance matrix in equation \eqref{eq:rbf} was $6.27\pm0.21\%$ for $M=1500$ which is substantially worse than the performance of the full GPC. As reference, it has been shown that the human error rate is approximately $2.5\%$.

Next we want to evaluate the two approximations to the marginal likelihood proposed in Section \ref{sec:ML}. We proceed by computing the accurate but expensive approximation $Z_{\rm ACC}$, the less accurate but affordable $Z_{\rm APP}$ and the marginal likelihood of the full GPC and the active set, simply denoted as $Z_{\rm EP}$ and $Z_{\rm EP,A}$, respectively. In order to show how the approximations depend on the size of the active set, we compute them for $p_{\rm inc}=\{0.5,0.6,\ldots,0.9,0.99,1,\}$, with $p_{\rm inc}=1$ being the full GPC. Figure \ref{fg:mla} shows that the three approximations approach the marginal likelihood of the full GPC as the inclusion threshold and so the active set size increases. As expected, $Z_{\rm ACC}$ is the best approximation, however the computational effort needed to compute it is roughly two orders of magnitude larger compared to the cost of computing $Z_{\rm APP}$ and $Z_{\rm EP,A}$. It is very interesting that even with large values of $p_{\rm inc}=0.99$ the size of the active set remains below $10\%$ of the training data and the contribution to the log-marginal likelihood from the inactive $Z_\mathrm{EP}(\btheta,\X,\y_I,\m_{I|A})$ set basically vanishes, since $Z_{\rm APP}$ and $Z_{\rm EP,A}$ are essentially the same.

\begin{figure}[!ht]
	\centering
		\centering
		\begin{psfrags}
			\psfrag{logd}[c][c][0.6][0]{Log-density}
			\psfrag{pos}[c][c][0.6][0]{$q(y^\star|\X,\y,\x^\star)$}
			\psfrag{pass-gp}[c][c][0.4][0]{PASS-GP}
			\psfrag{fpass-gp}[c][c][0.4][0]{fPASS-GP}
			\psfrag{ivm}[c][c][0.4][0]{\hspace{1mm} IVM}
			\psfrag{fullGPC}[c][c][0.4][0]{\hspace{0.5mm} full GPC}
			\subfigure[]{\includegraphics[scale=0.32]{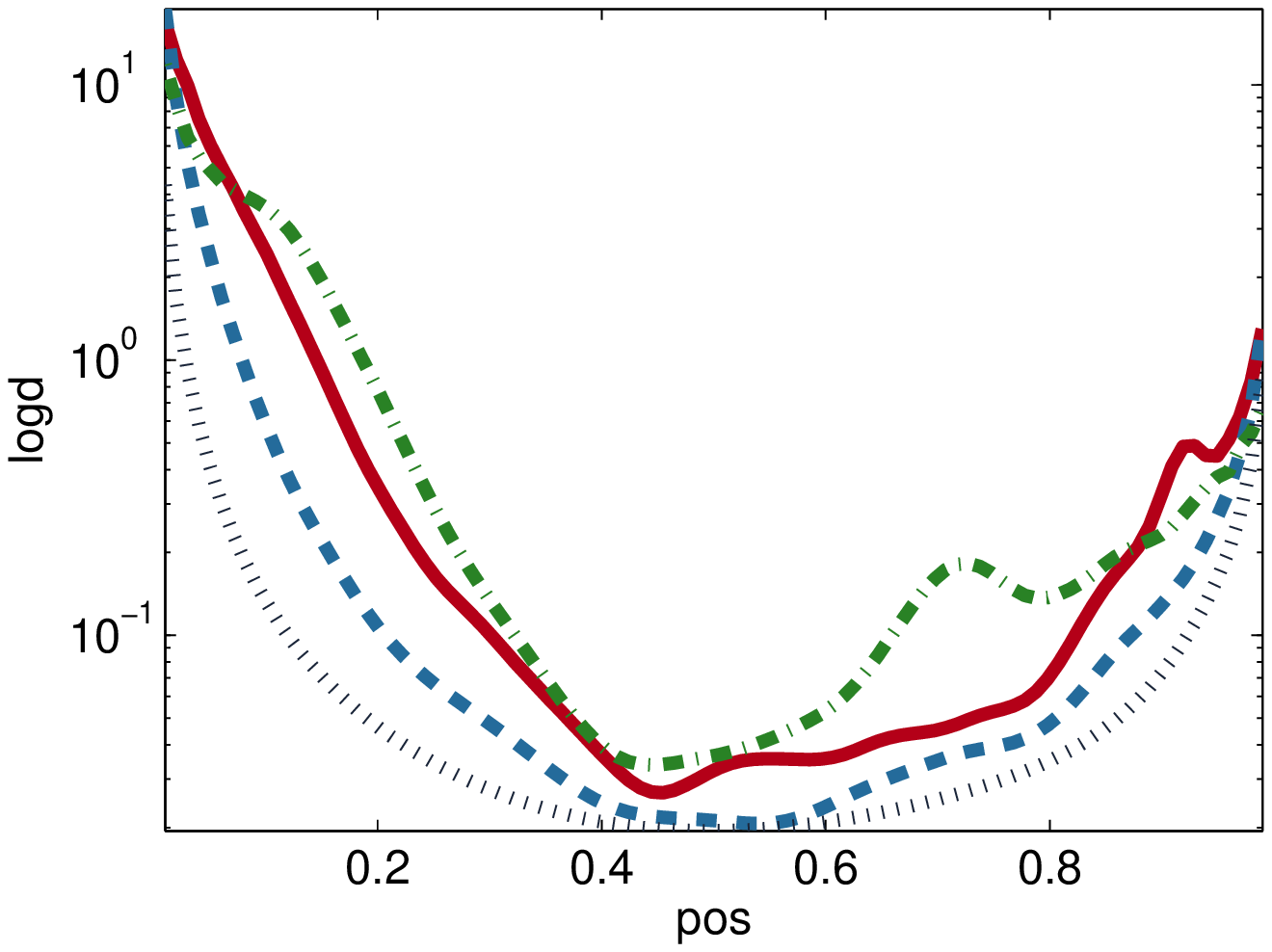}\label{fg:posdl}}
			\subfigure[]{\includegraphics[scale=0.32]{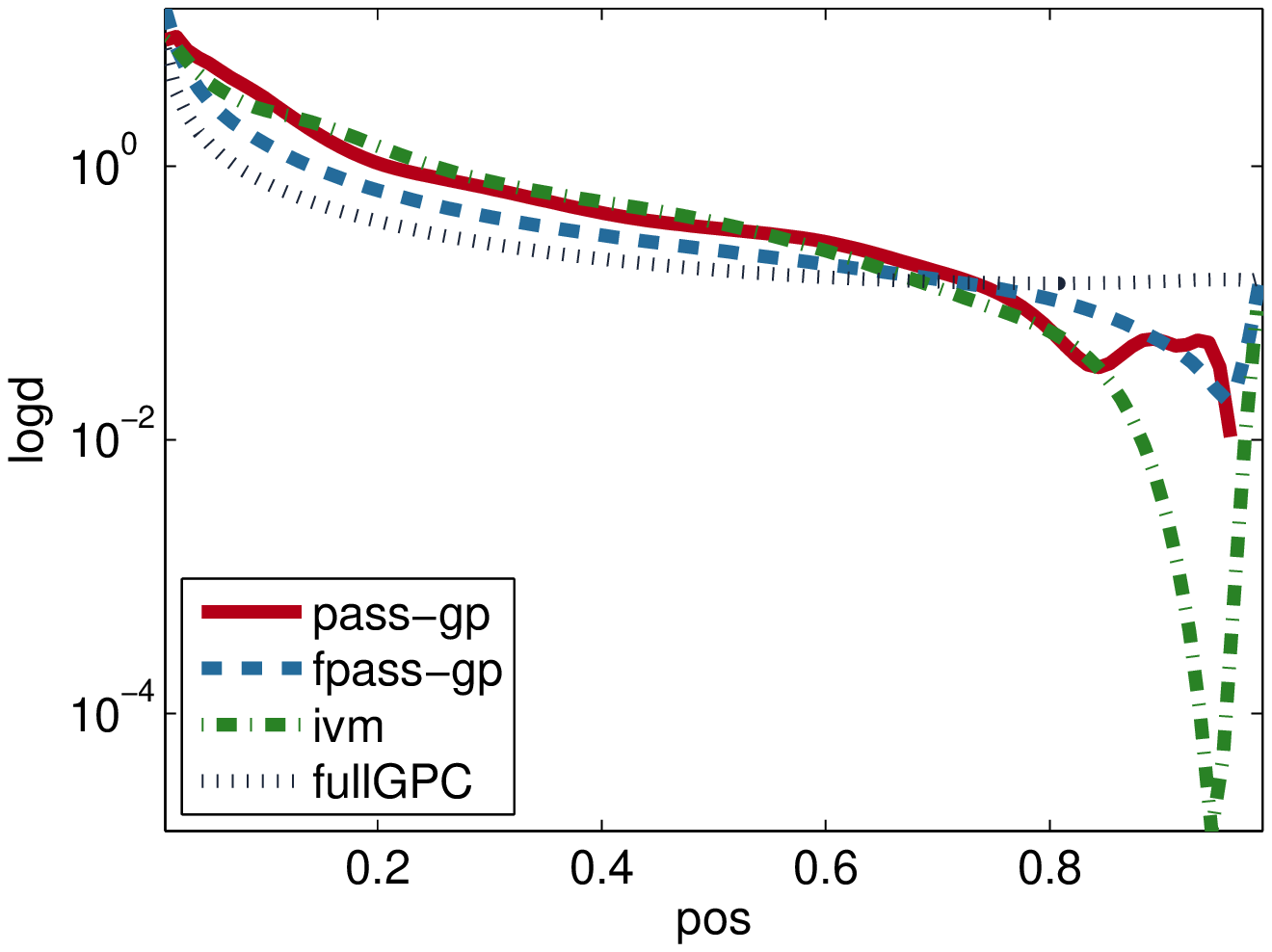}\label{fg:posdr}}
		\end{psfrags}
	\caption{Predictive density estimation for USPS data. Densities for correct and incorrect predictions are shown separately in (a) and (b), respectively. The ground truth for (a) is a two point mass mixture at $\{0,1\}$ and a flat distribution for (b).}
	\label{fg:posd}
\end{figure}

Finally, we want to asses the uncertainty of the predictions made by the Gaussian process based methods by means of comparing the predictive probabilities with the true outcomes. Figure \ref{fg:posd} shows estimated log predictive densities for PASS-GP, fPASS-GP, IVM and the full GPC, using all USPS predictions made on the test separated into correct and incorrect predictions. Assuming no labeling errors, the true density consists of two point mass densities at $\{0,1\}$ provided our one-against-rest setting. As one might expect, the full GPC achieves the best approximation, followed by fPASS-GP and PASS-GP. IVM suggests more predictive uncertainty because of the two \quotes{spurious} modes in Figure \ref{fg:posdl}. Another way to asses the predictive uncertainty is to compute Brier scores, that measures the average of square deviations between estimated and true predictive probabilities. For the USPS dataset we obtained: $0.53\pm0.03$, $0.27\pm0.01$, $0.71\pm0.02$ and $0.14\pm0.00$ for PASS-GP, fPASS-GP, IVM and full GPC, respectively. Note that the Brier scores are in agreement to what we observe in Figure \ref{fg:posd}.

\begin{figure}[!ht]
	\centering
		\centering
		\begin{psfrags}
			\psfrag{dig}[c][c][0.6][0]{Digit}
			\psfrag{act}[c][c][0.6][0]{Active set size}
			\psfrag{err}[c][c][0.6][0]{Error ($\%$)}
			\psfrag{error}[c][c][0.6][0]{Error ($\%$)}
			\psfrag{time}[c][c][0.6][0]{Time (s)}
			\psfrag{pass-gp}[c][c][0.5][0]{PASS-GP}
			\psfrag{fpass-gp}[c][c][0.5][0]{fPASS-GP}
			\psfrag{rsvm}[c][c][0.5][0]{\hspace{1mm} RSVM}
			\psfrag{ivm}[c][c][0.5][0]{\hspace{1mm} IVM}
			\subfigure[]{\includegraphics[scale=0.4]{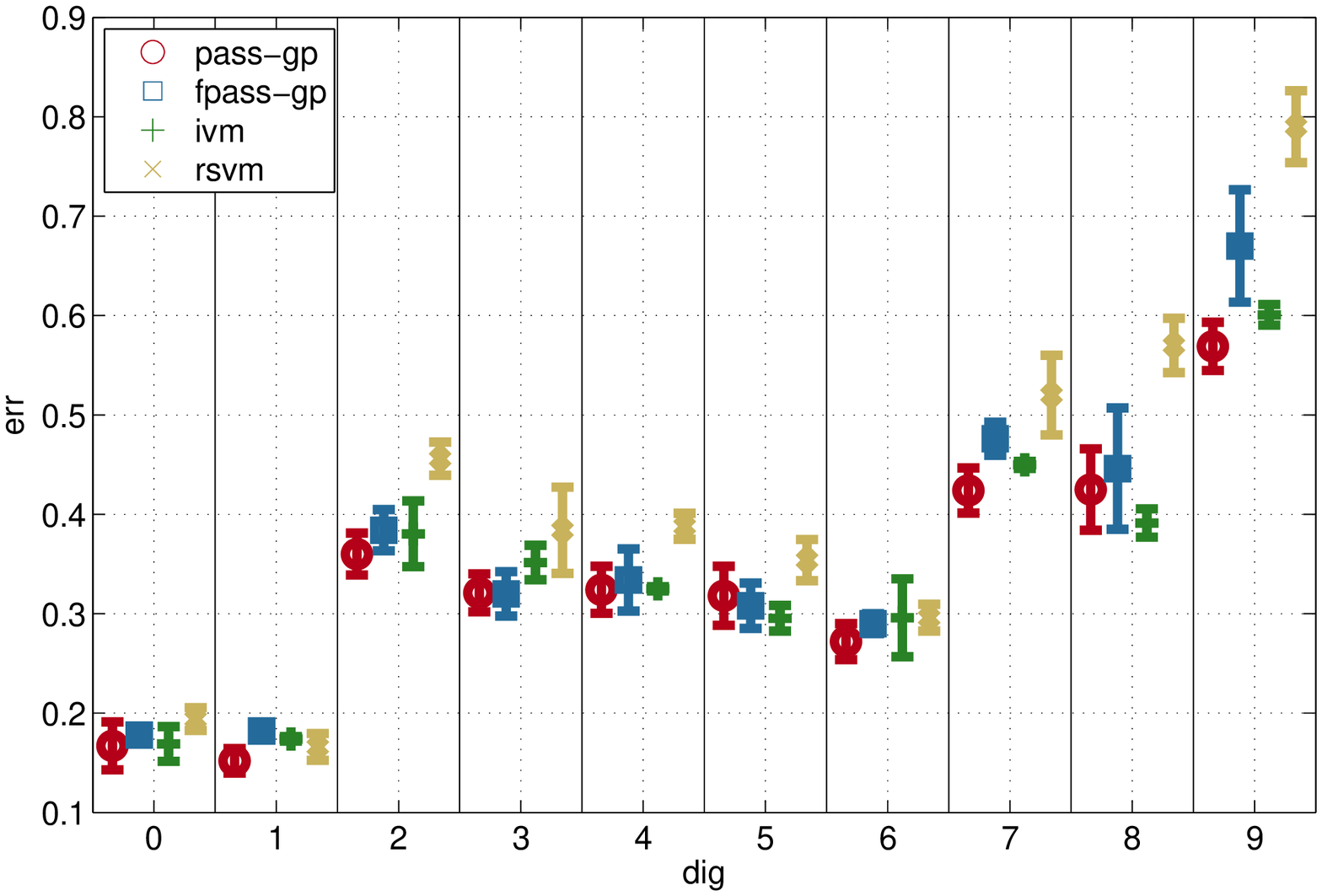}\label{fg:barsm}}
			\subfigure[]{\includegraphics[scale=0.4]{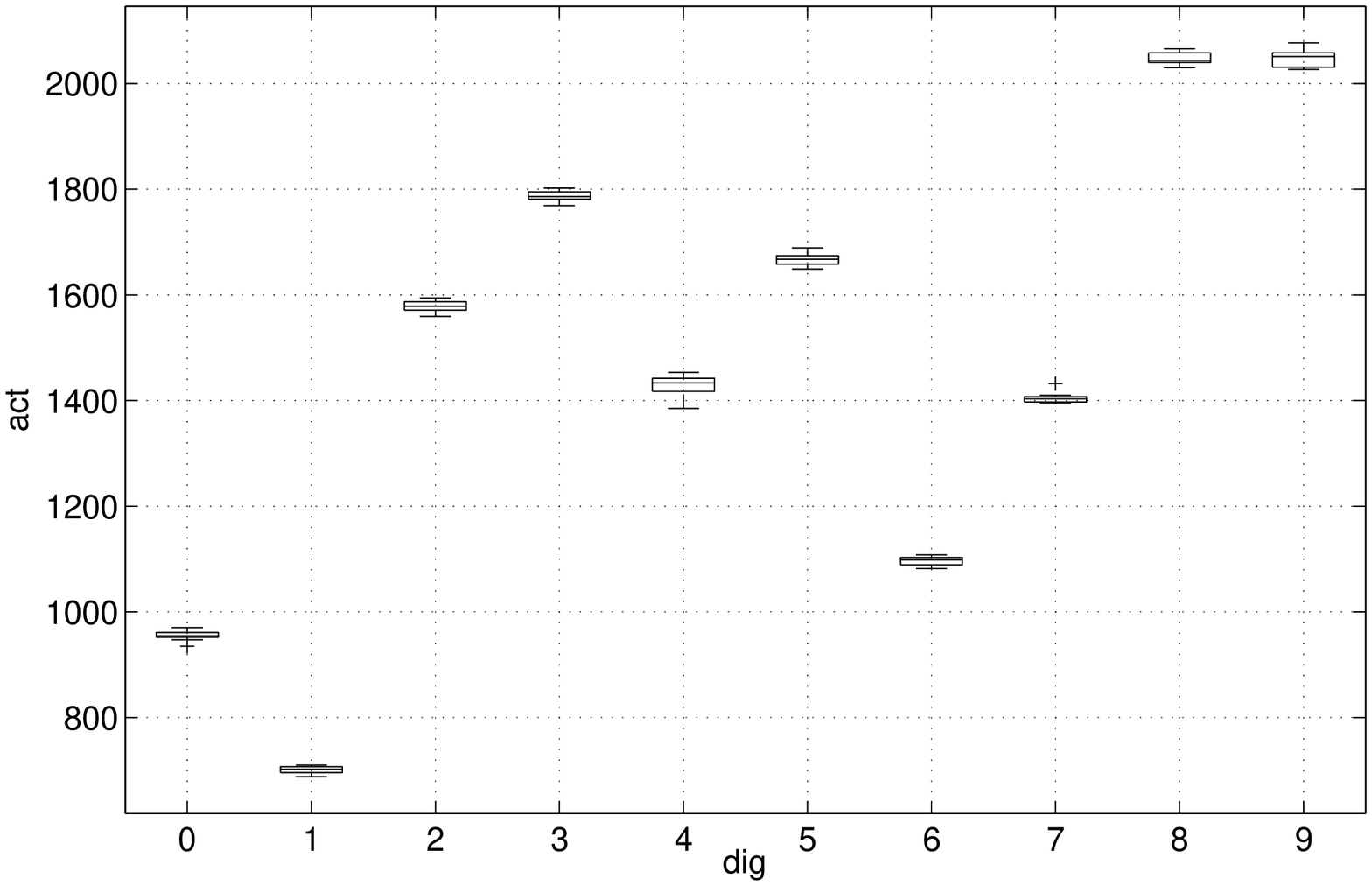}\label{fg:boxesm}}
			\subfigure[]{\includegraphics[scale=0.38]{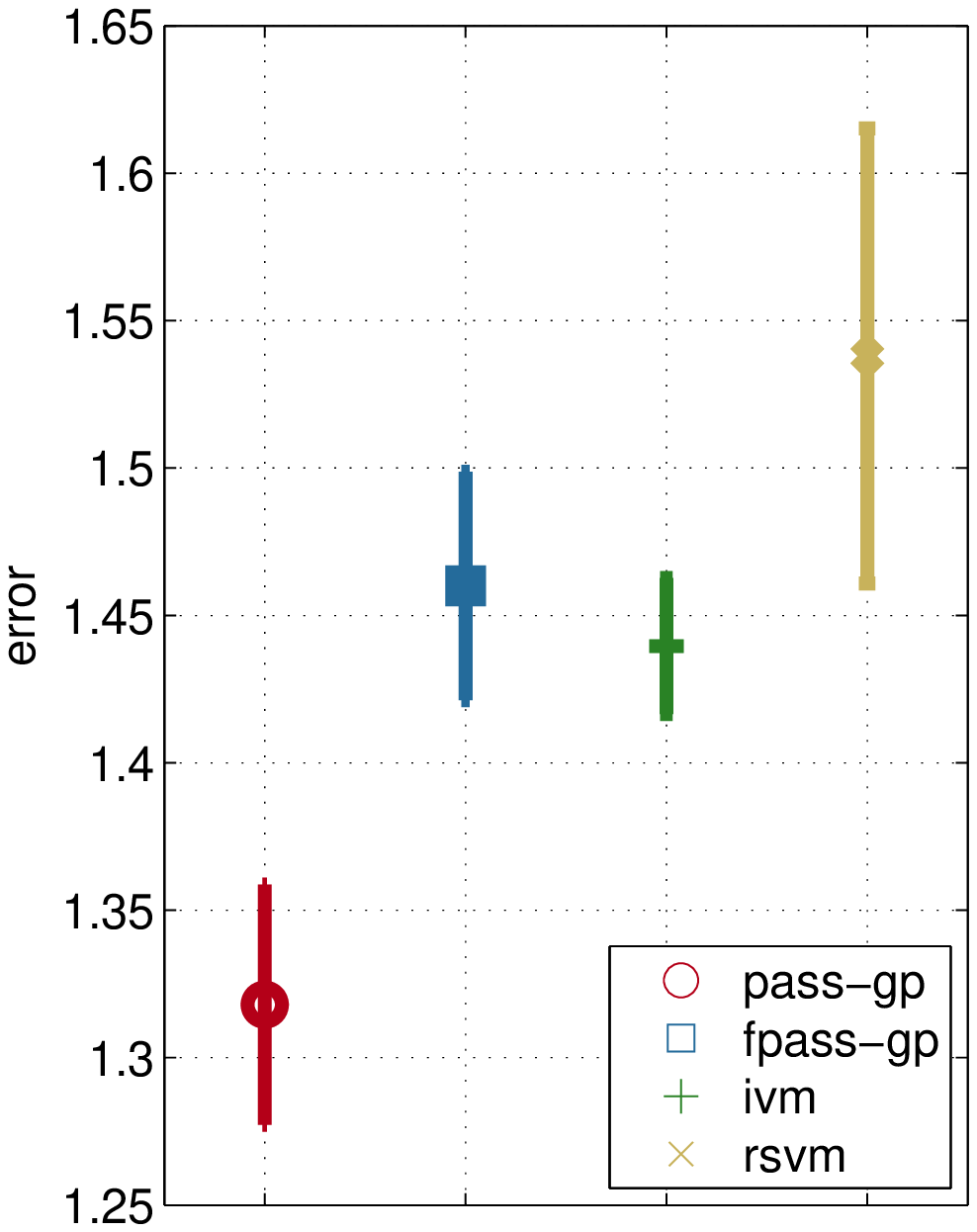}\label{fg:merrm}}
			\subfigure[]{\includegraphics[scale=0.38]{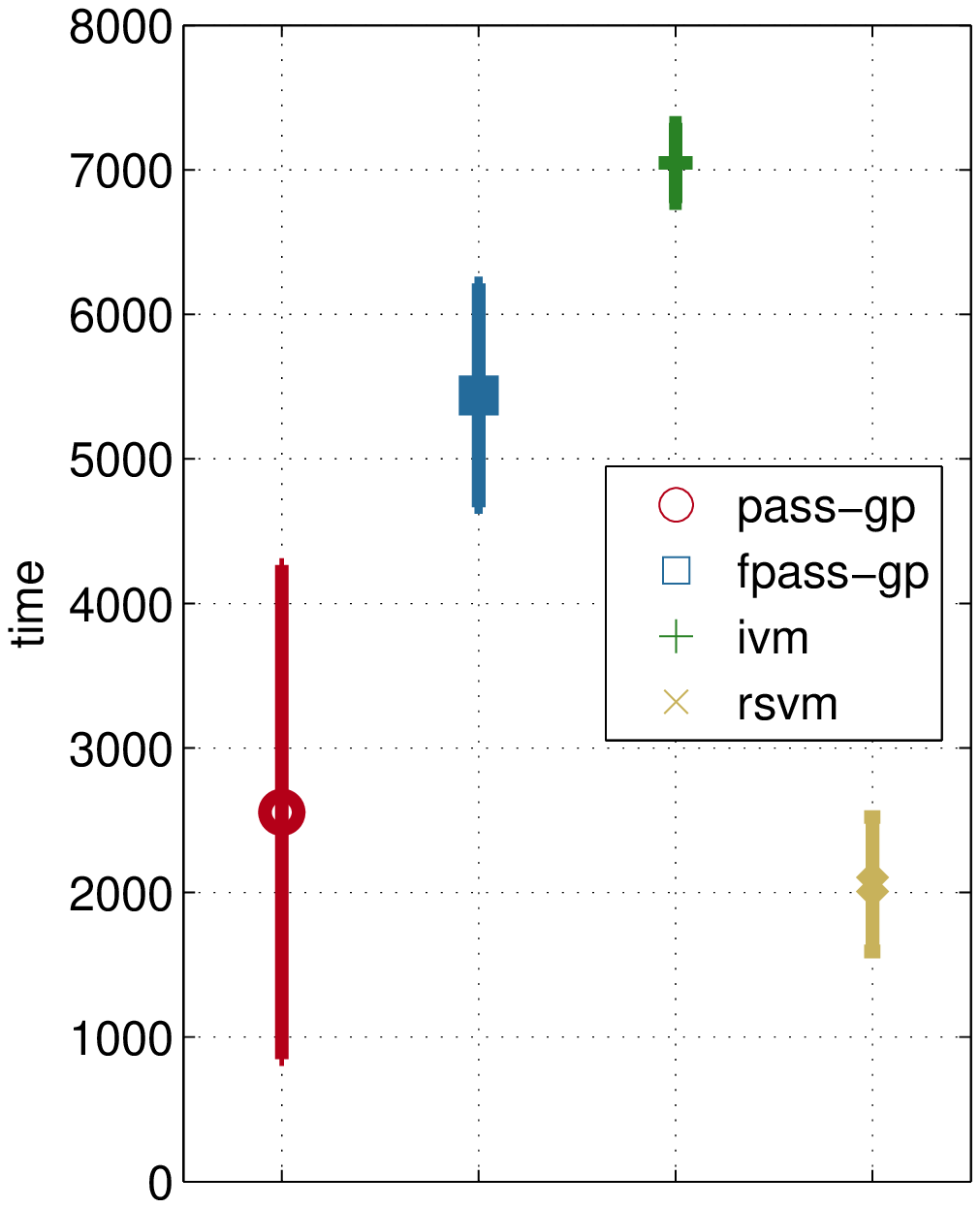}\label{fg:timem}}
		\end{psfrags}
	\caption{Error rates, active set sizes and run times for MNIST data. (a) Mean classification errors for each digit task using PASS-GP, fPASS-GP, RSVM and IVM. (b) Active set sizes for PASS-GP. Note that fPASS-GP, RSVM and IVM use $M=2000$. (c) Mean multi-class classification errors and (d) average timings over one-against-the-rest classifiers and repetitions. Error bars in (a), (c) and (d) are standard deviations computed over 10 repetitions of the experiment.}
	\label{fg:mnist}
\end{figure}

\subsection{MNIST}
The MNIST digits database has 60000 and 10000 as training and testing examples respectively. Each example is a gray-scale image of $28\times 28$ pixels. The estimated human test error is around $0.2\%$. The settings used for the algorithm are nearly the same as those for USPS with only two differences. $N_\mathrm{sub}$ is set 100 since the training set in MNIST is almost ten times larger than USPS and we are not updating the hyperparameter in each iteration but every 10-th, in order to make the training process faster. We also ran our algorithm with hyperparameter updates every single iteration without any noticeable improvement in performance (results not shown). Figure \ref{fg:mnist} shows test error rates, active set sizes, multi-class errors and running times for each binary classifier based on PASS-GP, fPASS-GP and RSVM using a 9-th degree polynomial covariance function
\begin{align*}
	K(\x_i,\x_j)=\theta_1(\x_i\cdot\x_j + 1)^9 \ .
\end{align*}
We use this covariance matrix instead for the standard squared exponential from equation \eqref{eq:rbf}, because a polynomial covariance is well known for providing optimal results for the MNIST dataset \cite{decoste02}. Results for the squared exponential covariance function can be found in \cite{henao10a} and confirm that the polynomial covariance behave slightly better for this dataset. For IVM we could not make the polynomial covariance to work properly, thus we decided to use the equation \eqref{eq:rbf} plus a linear term like in the USPS experiment.

\begin{table*}[ht]
	\centering
	\begin{tabular}{lcccccccccc}
	\hline
	Digit & 0 & 1 & 2& 3 & 4 & 5  & 6 & 7 & 8 & 9 \\
	\hline
	\small{USPS ($\%$)} & 0.63 & 0.38 & 1.01 & 0.69 & 0.93 & 1.16 & 0.51 & 0.37 & 0.59 & 0.65 \\
	Active set & 870 & 442 & 1251 & 1316 & 1654 & 1425 & 1242 & 987 & 1532 & 1281 \\
	\hline
	\small{MNIST ($\%$)} & 0.14 & 0.14 & 0.24 & 0.24 & 0.29 & 0.22 & 0.17 & 0.35 & 0.29 & 0.35 \\
	Active set & 6505 & 4372 & 11401 & 12988 & 9776 & 11960 & 7360 & 9872 & 15194 & 14790 \\
	\hline
	\end{tabular}
	\caption{Results for USPS and MNIST using PASS-GP and active set invariances. Figures are averages over 10 and 5 repetitions, respectively.}
	\label{tab:inv}
\end{table*}
From Figure \ref{fg:boxesm} it can be seen that in every case the size of the active set is less than $4\%$ of the training set. The results for fPASS-GP and RSVM were obtained using $M=2000$. We did try for larger values of $M$ but the reduction in error was not significant compared to the overhead in computational cost. Figure \ref{fg:barsm} shows the classification error for each digit. The performance of the three approaches considered is comparable but letting PASS-GP with an edge over the other two, both in terms of error and variances. Figure \ref{fg:merrm} shows the results of combining the ten binary classifiers. Again, PASS-GP behaves slightly better than the others, however when looking at the run times in Figure \ref{fg:timem} we can see that RSVM is computationally more affordable than our approaches, even more considering that it uses $M=2000$. Comparing PASS-GP to fPASS-GP, the former has a smaller mean run time but with larger variance compared to the more expensive fPASS-GP. fPASS-GP is more stable time-wise, but takes more time because it uses a fixed $M=2000$. 

As far as the authors know these are the first GP based results on MNIST using the whole database. IVM \cite{lawrence03} with sub-sampled images of size $13\times 13$ has been tried to produce a test error rate of $1.54\pm0.04\%$. Seeger \cite{seeger03} made additional tests on some digits (5, 8 and 9) on the full size images without any further improvement. On the other hand, PASS-GP is again comparable with state-of-the-art techniques not including preprocessing stages and/or data augmentation, for instance SVM is $1.4\%$ and $1.22\%$ using RBF and a 9-th degree polynomial kernel, respectively. The reported sizes of support vector sets are approximately two times larger than our active sets \cite{decoste02}.
\subsection{Incorporating Invariances}
It has been shown that a good way to improve the overall performance of a classifier is to incorporate additional prior knowledge in the training procedure particularly by means of externally handling invariances of the data. In \cite{decoste02}, it is shown that instead of just dealing with the invariances by augmenting the original dataset --- which turns out to be infeasible in many cases, it is better to augment only the support vector set of a SVM. We therefore try the same procedure as suggested in \cite{decoste02} consisting of four 1-pixel translations (left, up, right and down directions) on each element of the active set for USPS and eight 1-pixel translations (including diagonals as well) for MNIST, resulting in new training sets of size $5\times M$ and $9\times M$, accordingly. In this case we have used the same settings as in the previous experiments with only two differences. First, the hyperparameters have been set to those found using the original dataset. Second, we made the important observation that in order to get a performance improvement a large active set was needed. For training on the augmented dataset we increased $p_{\rm inc}$ from 0.6 to 0.99 for USPS and 0.9 for MNIST. We conjecture that we can get even better performance --- at the expense of a substantial increase in complexity, by increasing $p_{\rm inc}$ in the initial run to get a larger initial active set to work with.

Results in Table \ref{tab:inv} show that performance-wise, PASS-GP reached $3.35\pm0.03\%$ for USPS and $0.86\pm0.02\%$ for MINST on the multi-class task, what is comparable to state-of-the-art techniques. For instance SVM obtained $3.2\%$ on USPS and $0.68\%$ on MNIST with an equivalent procedure. The difference in performance is probably due to our active set not being large enough, since support set sizes reported for SVMs are typically twice as large \cite{decoste02}.
\subsection{IJCNN}
\begin{figure}[ht]
	\centering
		\centering
		\begin{psfrags}
			\psfrag{dig}[c][c][0.6][0]{Digit}
			\psfrag{nact}[c][c][0.6][0]{Active set size}
			\psfrag{err}[c][c][0.6][0]{Error ($\%$)}
			\psfrag{time}[c][c][0.6][0]{Time (s)}
			\psfrag{fpass-gp}[c][c][0.5][0]{fPASS-GP}
			\psfrag{rsvm}[c][c][0.5][0]{\hspace{1mm} RSVM}
			\psfrag{ivm}[c][c][0.5][0]{\hspace{1mm} IVM}
			\subfigure[]{\includegraphics[scale=0.38]{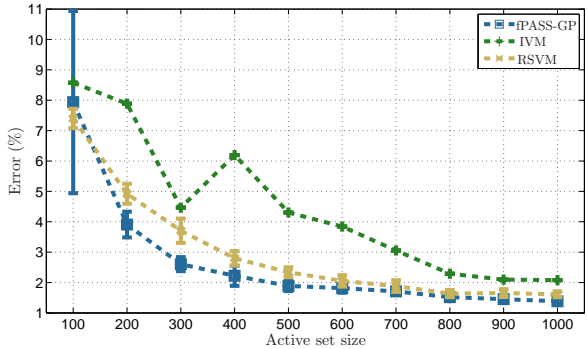}\label{fg:errnact}}
			\subfigure[]{\includegraphics[scale=0.38]{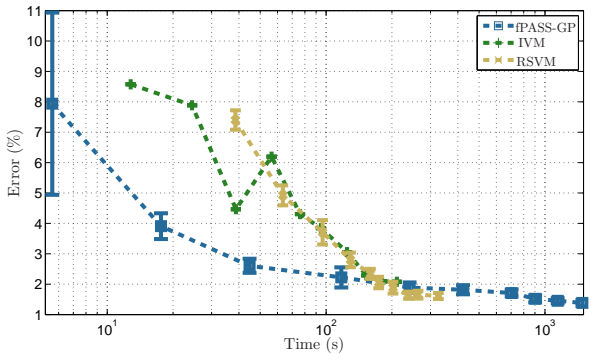}\label{fg:errtime}}
		\end{psfrags}
	\caption{Error rates and run times for IJCNN data. (a) Mean classification error as a function of the active set size using fPASS-GP, RSVM and IVM. (b) Mean classification error as a function of the run time. Error bars correspond to standard deviations computed over 10 repetitions of the experiment.}
	\label{fg:ijcnn}
\end{figure}

As final experiment, we want to compare fPASS-GP, RSVM and IVM on a common ground. For this purpose we use the IJCNN dataset which is widely used by the SVM research community. It consists of 49990 training examples, 91701 test examples and each observation counts with 22 features. We consider $M=\{100,200,\ldots,1000\}$ with squared covariance function and fixed hyperparameters, the latter using the values suggested in \cite{keerthi06}, that is $\btheta=[1 \ 1/8 \ 1/16]$ for f-PASSGP and IVM, and $\btheta=[1 \ 1/8 \ 0]$, $C=16$ for RSVM. For IVM we include a linear term as in the previous experiments with $\theta_4=1$. Besides, each setting was repeated 10 times to collect statistics. Figure \ref{fg:ijcnn} summarizes the results obtained. More specifically, Figure \ref{fg:errnact} shows the mean classification error as a function of the active set. We can see that fPASS-GP is slightly better than RSVM and IVM in the entire range of $M$, besides the former seems to be particularly good for small values of $M$. When we plot mean errors as a function of running times --- as a proxy for the computational cost, we see that there exist two regimes, one for small values of $M$ where fPASS-GP outperforms RSVM and IVM, and the other where the cubic complexity of the GPCs start hurting fPASS-GP, thus letting RVM and IVM with a better error-cost trade-off.
\section{Discussion} \label{sec:disc}
We have proposed a framework for active set selection in GPC. The core of our active set update rule is that the predictive distribution of a GPC can be used to quantify the relative weight of points in the active set that can be marked for deletion or new points from the active set with low predictive probabilities, that make them ideal for inclusion. The algorithmic skeleton of our framework consists on two alternating steps, namely active set updates and hyperparameter optimization. We designed two active set update criteria that target two different practical scenarios. The first we called PASS-GP focuses on interpretability of the parameters of the update rule by thresholding the predictive distributions of GPC. The second acknowledges that in some applications having a fixed computational cost is key, thus fPASS-GP keeps the size of the active set fixed so the overall cost and memory requirements can be known beforehand. 

We presented theoretical and practical support that our active set selection strategy is efficient while still retaining the most appealing benefits of GPC: prediction uncertainty, model selection, prior knowledge leverage and state-of-the-art performance. Compared to other approximative methods, although slower than IVM \cite{lawrence03} and RSVM \cite{keerthi06}, PASS-GP provides better results. We did not consider any auxiliary set method like FITC \cite{naish08} because for task of the size like for example MNIST or IJCNN, it is prohibitive. Additionally, we have noticed in practice that our approximation is quite insensitive to the initial active set selection and also that more than two or three passes through the data do not yield improved performance nor large active set sizes. The code used in this work is based on the Matlab toolbox provided with \cite{rasmussen06} and is publicly available at \url{http://cogsys.imm.dtu.dk/passgp}.

The not so satisfying feature of active set approximations, is that we are ignoring some of the training data. Although some of our findings on the USPS data set actually suggest that this can be beneficial for performance, it is of interest to make a modified version where the inactive set is used approximately in a cost efficient way. The representer theorem for the mean prediction and the approximations for marginal likelihood discussed in this paper might give inspiration for such methods. In conclusion, efficient methods for GPs are still much in need when the data is abundant such as in ordinal regression for collaborative filtering.
\bibliographystyle{elsarticle-num}
\bibliography{../bib_files/mlbib}
%
\end{document}